\documentclass[11pt]{article}

\usepackage[final]{acl}

\usepackage{times}
\usepackage{latexsym}

\usepackage[T1]{fontenc}
\usepackage[utf8]{inputenc}
\usepackage{microtype}
\usepackage{inconsolata}
\usepackage{graphicx}
\usepackage{booktabs}
\usepackage{multirow}
\usepackage{amsmath}

\title{PSK at SemEval-2026 Task 9: Multilingual Polarization Detection\\Using Ensemble Gemma Models with Synthetic Data Augmentation}

\author{Srikar Kashyap Pulipaka \\
  Independent Researcher \\
  \texttt{\{srikar.kashyap@gmail.com\}} \\}

\begin{document}
\maketitle

\begin{abstract}
We present our system for SemEval-2026 Task 9: Multilingual Polarization Detection, a binary classification task spanning 22 languages. Our approach fine-tunes separate Gemma~3 models (12B and 27B parameters) per language using Low-Rank Adaptation (LoRA), augmented with synthetic data generated by a large language model (LLM). We employ three synthetic data strategies (direct generation, paraphrasing, and contrastive pair creation) using GPT-4o-mini, with a multi-stage quality filtering pipeline including embedding-based deduplication. We find that per-language threshold tuning on the development set yields 2 to 4\% F1 improvements without retraining. We also use weighted ensembles of 12B and 27B model predictions with per-language strategy selection. Our final system achieves a mean macro-F1 of 0.811 across all 22 languages, ranking 2nd overall of the participating teams, with 1st place finishes in 3 languages and top-3 in 8 languages. We also find that alternative architectures (XLM-RoBERTa, Qwen3) that showed strong development set performance suffered 30 to 50\% F1 drops on the test set, highlighting the importance of generalization.
\end{abstract}

\section{Introduction}

Team PSK participated in SemEval-2026 Task 9 \citep{naseem-etal-2026-polar} on detecting attitude polarization in social media text across 22 typologically diverse languages. The task requires binary classification of social media text as polarized or non-polarized, where polarization encompasses stereotyping, vilification, dehumanization, and intolerance.

The task presents several challenges: the dataset sizes vary across languages, polarization manifests differently across cultures, and some languages exhibit class imbalance. In our work, we explore per-language fine-tuned Gemma models with synthetic data augmentation, ensemble methods, and post-hoc threshold tuning. We find that robust generalization, rather than maximizing development set performance, is the critical factor for success.

The remainder of this paper is structured as follows: Section~\ref{sec:related} gives a brief overview of related work, Section~\ref{sec:data} describes the data, Section~\ref{sec:methodology} describes our methodology, Section~\ref{sec:results} discusses our results, and we conclude in Section~\ref{sec:conclusion}.

\section{Related Work}
\label{sec:related}

Recent work has explored LLM-generated data for text classification augmentation. \citet{cegin2025llms} found that LLM augmentation primarily benefits low-data scenarios. \citet{yong2024lexcgen} demonstrated 5.6 to 8.9 point improvements for extremely low-resource languages using lexicon-conditioned generation.
Our augmentation pipeline draws on backtranslation \citep{sennrich2016improving}. \citet{edunov2018understanding} demonstrated that noisy backtranslations provide stronger training signals than clean outputs. Our contrastive pair generation is inspired by counterfactual data augmentation for hate speech detection \citep{mostafazadeh-davani-etal-2021-improving}, where minimal-edit counterfactuals improve classifier robustness.

We use LoRA \citep{hu2021lora} for parameter-efficient fine-tuning, and our per-language threshold tuning follows work by \citet{lipton2014thresholding} on F1 maximization and \citet{pillai2013threshold} on per-class threshold optimization for macro-averaged metrics.

\section{Data}
\label{sec:data}

\subsection{Task Data}

SemEval-2026 Task 9 Subtask 1 uses the POLAR dataset \citep{naseem2026polarbenchmarkmultilingualmulticultural}, a binary classification task: given a social media text in one of 22 languages, the task is to classify it as polarized (1) or non-polarized (0). The primary evaluation metric is macro-averaged F1-score. The 22 languages are: Amharic (amh), Arabic (arb), Bengali (ben), Burmese (mya), Chinese (zho), English (eng), German (deu), Hausa (hau), Hindi (hin), Italian (ita), Khmer (khm), Nepali (nep), Odia (ori), Persian (fas), Polish (pol), Punjabi (pan), Russian (rus), Spanish (spa), Swahili (swa), Telugu (tel), Turkish (tur), and Urdu (urd). Training set sizes range from approximately 1,700 (Punjabi) to 7,000 (Swahili) samples. Several languages exhibit class imbalance, most notably Khmer (10:1 polarized), Hausa (8:1 non-polarized), Hindi (6:1 polarized), and Amharic (3:1 polarized).

We split the original data into 80\% train and 20\% validation before adding any synthetic data, ensuring the validation set contains only real data.

\subsection{Data Augmentation}
\label{sec:augmentation}

To provide more data to our models, we generated synthetic training data using GPT-4o-mini with three complementary strategies:

\paragraph{Direct Generation (50\%).} We generated new samples natively in the target language, covering five culturally relevant topic categories: political, ethnic/racial, religious, social class, and international relations.

\paragraph{Label-Preserving Paraphrasing (30\%).} We created paraphrases of real training samples with temperature 0.7, filtered to ensure cosine similarity below 0.90 with the original.

\paragraph{Contrastive Pairs (20\%).} We generated minimal pairs on the same topic, one polarized and one non-polarized, to sharpen class boundaries.

In preliminary experiments for lower-resource languages, we also explored backtranslation through pivot languages and cross-lingual transfer from related languages (e.g., Hindi to Nepali, Bengali to Odia).

\paragraph{Quality Filtering.} We applied a multi-stage pipeline to ensure data quality: (1) basic cleaning and length filtering, (2) label leakage detection via regex patterns, (3) embedding-based deduplication using \texttt{paraphrase-multilingual-MiniLM-L12-v2} \citep{reimers2019sentence} with a 0.90 cosine similarity threshold (both intra-synthetic and against original training data), and (4) round-trip translation consistency checks (threshold 0.70) for translated samples.

Synthetic samples were added only to the training portion at a configurable ratio. We generated approximately 1,000 synthetic samples per language.

\section{Methodology}
\label{sec:methodology}

\begin{figure*}[t]
\centering
\includegraphics[width=\textwidth]{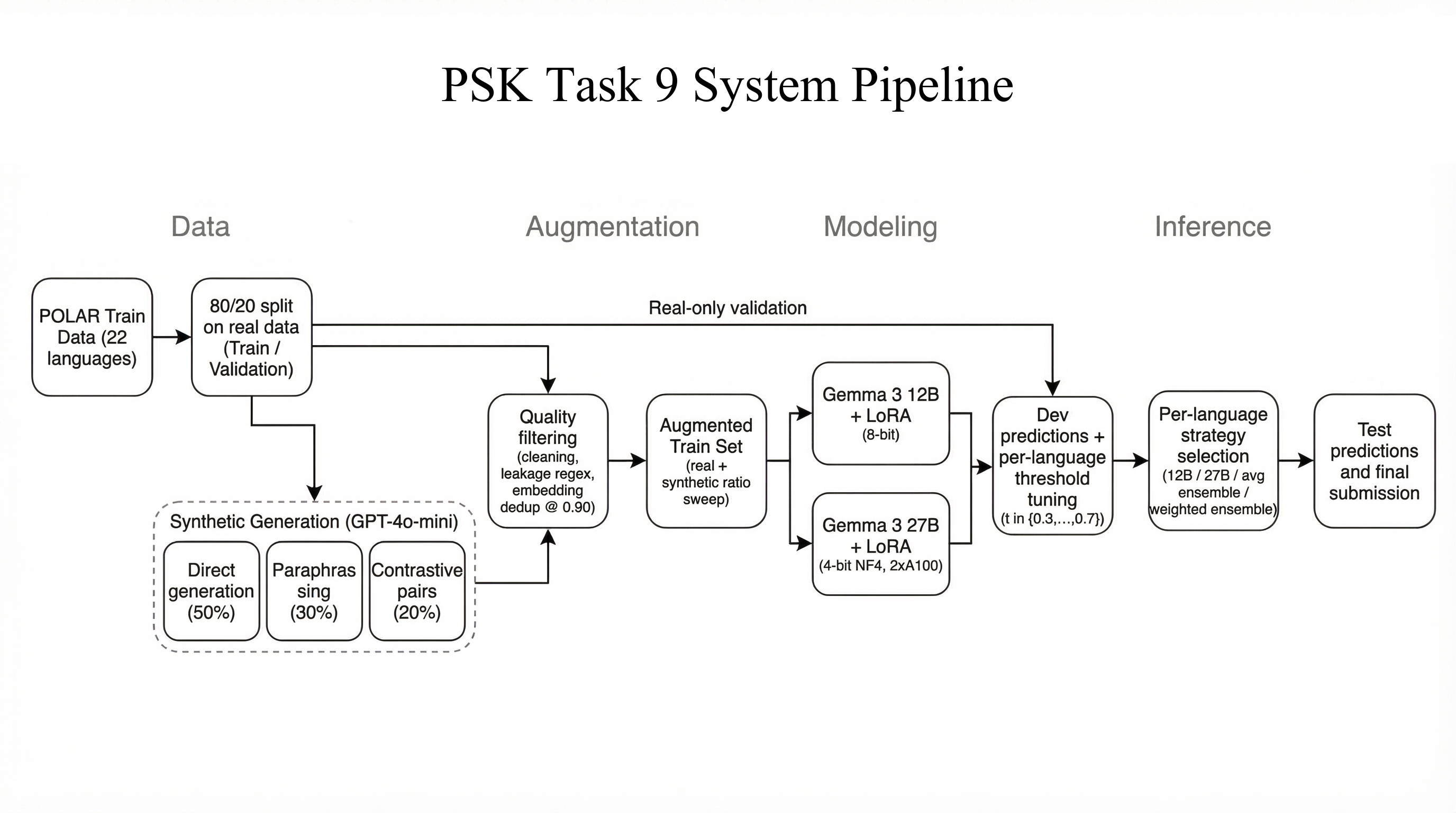}
\caption{Overview of our multilingual polarization detection pipeline, from real-data splitting and synthetic augmentation to Gemma model training, threshold tuning, and per-language strategy selection.}
\label{fig:pipeline}
\end{figure*}

Figure~\ref{fig:pipeline} provides an overview of our end-to-end system pipeline.

\subsection{Model Architecture}

We chose Google's Gemma~3 family \citep{gemma2024} as our primary model because of its broad multilingual pretraining mix, which provides meaningful coverage across all 22 POLAR languages rather than concentrating on a narrow set of high resource ones. Our model comparison in Section~\ref{sec:results} confirms this choice: Gemma transferred from development to test substantially better than XLM-RoBERTa or Qwen3, both of which have more skewed pretraining distributions. We fine-tuned the Gemma~3 family using LoRA with a sequence classification head. We used two model sizes:

\begin{itemize}
    \item \textbf{Gemma~3 12B}: Our primary model. We used 8-bit quantization on a single A100 GPU, with LoRA rank 16, a learning rate of $5 \times 10^{-5}$, and trained for 3 epochs.
    \item \textbf{Gemma~3 27B}: Our ensemble model. We used 4-bit NF4 quantization \citep{dettmers2023qlora} with gradient checkpointing across 2 A100 GPUs via \texttt{device\_map="auto"}.
\end{itemize}

Both models use 2 output labels, eager attention, and \texttt{bfloat16} precision. We set the pad token to the end-of-sequence (EOS) token. All models were trained with batch size 1 to 2, gradient accumulation of 2 to 4 steps (effective batch size 4), warmup ratio 0.1, max sequence length 128 to 256 tokens, and random seed 42. We used the AdamW optimizer with cosine learning rate decay.

\subsection{Ensemble and Threshold Tuning}
\label{sec:ensemble}

\paragraph{Threshold Tuning.} For each language, we searched over thresholds $t \in \{0.3, 0.35, 0.4, \ldots, 0.7\}$ to maximize development set macro-F1, rather than using the default $t = 0.5$.

\paragraph{Ensemble Strategies.} For each language, we evaluated four strategies on the development set:
\begin{enumerate}
    \item 12B predictions with tuned threshold
    \item 27B predictions with tuned threshold
    \item Average ensemble: $p = \frac{p_{\text{12B}} + p_{\text{27B}}}{2}$
    \item Weighted ensemble: $p = w \cdot p_{\text{12B}} + (1{-}w) \cdot p_{\text{27B}}$, with $w \in \{0.3, 0.4, 0.6, 0.7\}$
\end{enumerate}
We selected the strategy achieving the highest development F1 per language.

\subsection{Infrastructure}

Experiments were run on a computing cluster with NVIDIA A100 40GB GPUs. The 12B model requires 1 GPU while the 27B model requires 2 GPUs.

\section{Results}
\label{sec:results}

We first provide an overview of our official results, then discuss the results of our internal experiments on the development data.

\subsection{Official Results}

We were allowed to submit 5 systems. Table~\ref{tab:submissions} summarizes our test submissions.

\begin{table}[t]
\centering
\small
\begin{tabular}{clc}
\toprule
\textbf{Sub} & \textbf{Strategy} & \textbf{Mean F1} \\
\midrule
1 & Best Gemma-12B per lang & 0.797 \\
2 & Hybrid (XLM-R, Qwen3) & 0.665 \\
4 & Ensemble + threshold & 0.811 \\
\textbf{5} & \textbf{Best of Sub 1 + Sub 4} & \textbf{0.812} \\
\bottomrule
\end{tabular}
\caption{Test submission results. Submission 2 used alternative architectures that failed to generalize. The final submission selects the best per-language results across submissions.}
\label{tab:submissions}
\end{table}

Our best official submission achieved a mean macro-F1 of 0.811 across all 22 languages. A post-hoc combined run that selected the best per-language result between Submission~1 (12B only) and Submission~4 (ensemble with threshold tuning) reached 0.812. Submission~2, which used XLM-RoBERTa and Qwen3 for selected languages, performed significantly worse with a mean F1 of 0.665.

\subsection{Synthetic Data Ablation}

Table~\ref{tab:synth_ablation} shows the impact of synthetic data ratio on mean development F1 across all languages.

\begin{table}[t]
\centering
\small
\begin{tabular}{lc}
\toprule
\textbf{Synth.\ Ratio} & \textbf{Dev F1} \\
\midrule
0\% (baseline) & 0.812 \\
10\% & 0.818 \\
20\% & 0.820 \\
\textbf{30\%} & \textbf{0.822} \\
50\% & 0.819 \\
\bottomrule
\end{tabular}
\caption{Mean development F1 for Gemma~3 12B by synthetic data ratio. A ratio of 30\% is optimal.}
\label{tab:synth_ablation}
\end{table}

We find that lower-resource languages benefit most from synthetic augmentation (+2.2\%), while higher-resource languages see smaller gains (+1.9\%). Some languages are still sensitive to synthetic data choices: Swahili improves at low ratios but does not benefit from higher ratios, and Khmer degrades by 6.4\% at 50\% synthetic ratio. This is consistent with the finding by \citet{cegin2025llms} that augmentation primarily benefits lower-data scenarios.

\subsection{Model Comparison}

We evaluated multiple architectures. Table~\ref{tab:model_comparison} shows the results.

\begin{table}[t]
\centering
\small
\begin{tabular}{lccc}
\toprule
\textbf{Model} & \textbf{Dev F1} & \textbf{Test F1} & \textbf{$\Delta$} \\
\midrule
Gemma~3 12B & 0.827 & 0.797 & $-$0.030 \\
Gemma~3 27B & 0.838 & -- & -- \\
XLM-R Large & 0.800 & 0.665\textsuperscript{*} & $-$0.135 \\
Qwen3-14B & 0.811\textsuperscript{\dag} & 0.665\textsuperscript{*} & $-$0.146 \\
\bottomrule
\end{tabular}
\caption{Model comparison. \textsuperscript{*}Hybrid submission. \textsuperscript{\dag}Best dev F1 for selected languages only. Only Gemma models generalize reliably from dev to test.}
\label{tab:model_comparison}
\end{table}

Only the Gemma models generalize reliably from development to test. XLM-RoBERTa and Qwen3 showed competitive development performance but suffered large drops on the test set. For example, XLM-R achieved 0.856 development F1 for Bengali but only 0.297 on test, a 56\% absolute drop. This trend is similar to what we observed in our previous work \citep{mhalgi2024iucl}, where ensembles that performed best on the development set did not fully generalize to the test set.

\subsection{Ensemble and Threshold Results}

Table~\ref{tab:strategies} shows the distribution of winning strategies across languages.

\begin{table}[t]
\centering
\small
\begin{tabular}{lc}
\toprule
\textbf{Strategy} & \textbf{Languages Won} \\
\midrule
Ensemble weighted & 9 \\
Ensemble average & 5 \\
12B tuned only & 4 \\
27B tuned only & 4 \\
\bottomrule
\end{tabular}
\caption{Best strategy per language on the development set. Ensemble methods win for 14 out of 22 languages.}
\label{tab:strategies}
\end{table}

Ensemble methods won for 14 out of 22 languages. Optimal thresholds range from 0.3 (Nepali) to 0.7 (Khmer, Amharic, Bengali), showing that the models are not well calibrated at the default threshold of 0.5.

Table~\ref{tab:perlang} shows per-language test F1 for our key submissions. Overall, 18 out of 22 languages improved from Submission~1 to Submission~4, with the largest improvement for Khmer (+8.7\%).

\begin{table}[t]
\centering
\small
\begin{tabular}{lccr}
\toprule
\textbf{Lang} & \textbf{Sub 1} & \textbf{Sub 4} & \textbf{$\Delta$} \\
\midrule
khm & 0.656 & 0.743 & +8.7\% \\
swa & 0.774 & 0.811 & +3.7\% \\
nep & 0.876 & 0.908 & +3.2\% \\
fas & 0.804 & 0.828 & +2.4\% \\
hin & 0.800 & 0.824 & +2.4\% \\
pol & 0.814 & 0.835 & +2.1\% \\
ita & 0.543 & 0.563 & +2.0\% \\
ori & 0.793 & 0.811 & +1.8\% \\
eng & 0.805 & 0.818 & +1.3\% \\
ben & 0.828 & 0.837 & +0.9\% \\
\midrule
\multicolumn{2}{l}{\textbf{Mean (all 22)}} & \textbf{0.811} & \textbf{+1.4\%} \\
\bottomrule
\end{tabular}
\caption{Per-language test F1 improvements (top 10 shown). 18 out of 22 languages improved from Submission~1 to Submission~4.}
\label{tab:perlang}
\end{table}

\subsection{Analysis}

\paragraph{Threshold Sensitivity.} We find that the models exhibit probability miscalibration. For example, Russian predictions have a mean probability of 0.246 (under-confident), while Khmer averages 0.919 (over-confident). This miscalibration makes the default 0.5 threshold suboptimal, which explains why threshold tuning alone yields significant gains.

\paragraph{Leaderboard Performance.} On the official leaderboard, our system ranked 2nd overall out of 60 teams, with a mean F1 of 0.811 compared to 0.818 for the top-ranked system. We achieved 1st place in 3 languages (Amharic, Hindi, Swahili), top-3 in 8 languages, and top-10 in 17 out of 22 languages. Our weakest rankings are in Italian (25th) and Spanish (14th), which are also our lowest-performing languages in absolute terms. Full per-language rankings are provided in Appendix~\ref{sec:leaderboard}.

\paragraph{Remaining Challenges.} Italian (0.563) is our worst performing language by a wide margin. On closer inspection, the POLAR Italian split reveals a topic coverage gap rather than a simple class ratio shift. The training and development sets contain zero examples labeled with the \texttt{political} or \texttt{other} topic categories, yet these two categories account for 41\% of the Italian test set (631 of 1538 samples) and roughly 87\% of its positive class, as shown in Table~\ref{tab:ita_shift}. Since POLAR topic labels only co-occur with polarized rows, the model was effectively evaluated on a positive sub population it never observed during training.

\begin{table}[t]
\centering
\small
\begin{tabular}{lrrr}
\toprule
\textbf{Topic} & \textbf{Train} & \textbf{Dev} & \textbf{Test} \\
\midrule
political       & 0 (0.0\%)    & 0 (0.0\%)    & 412 (26.8\%) \\
racial/ethnic   & 746 (22.4\%) & 37 (22.3\%)  & 143 (9.3\%)  \\
religious       & 285 (8.5\%)  & 14 (8.4\%)   & 69 (4.5\%)   \\
gender/sexual   & 381 (11.4\%) & 19 (11.4\%)  & 61 (4.0\%)   \\
other           & 0 (0.0\%)    & 0 (0.0\%)    & 219 (14.2\%) \\
\midrule
positive rate   & 41.0\%       & 41.6\%       & 47.3\%       \\
\bottomrule
\end{tabular}
\caption{Italian POLAR topic label and positive class distribution across splits. The \texttt{political} and \texttt{other} categories are entirely absent from train and development but account for 41\% of test samples and about 87\% of test positives, which explains the Italian test set performance drop.}
\label{tab:ita_shift}
\end{table}

\section{Conclusion}
\label{sec:conclusion}

We conducted an extensive analysis of Gemma-based models for multilingual polarization detection, investigating the effectiveness of synthetic data augmentation, ensemble methods, and per-language threshold optimization. We find that model generalization matters more than development set performance: Gemma was the only architecture that reliably transferred from development to test, while XLM-RoBERTa and Qwen3 suffered large drops. Threshold tuning provides 2 to 4\% improvement without retraining. Ensemble methods combining 12B and 27B predictions are effective for 14 out of 22 languages. Synthetic data provides modest gains of 2 to 3\%, primarily for low-resource languages, with 30\% being the optimal ratio. Our final system achieves 0.811 mean macro-F1 across 22 languages, placing 2nd overall on the official leaderboard.

For future work, we plan to explore more sophisticated calibration methods such as temperature scaling and investigate topic coverage gaps of the kind we observed in Italian, where an entire class of polarization topics was absent from the training data.

\section*{Limitations}

Our system trains separate models per language, requiring significant compute resources. The synthetic data quality for low-resource languages (Khmer, Swahili) may be limited, as GPT-4o-mini may not generate authentic text in these languages. Our threshold tuning is optimized on a small development set and may not fully transfer to the test distribution. Finally, the synthetic data mix of 50\% direct generation, 30\% paraphrasing, and 20\% contrastive pairs was chosen heuristically, based on the intuition that direct generation should contribute the bulk of new topical content, paraphrasing should preserve label boundaries on real samples, and contrastive pairs should sharpen class separation. We did not run a systematic ablation over these strategy ratios and leave that study to future work.

\section*{Ethics Statement}

Our system processes social media text that may contain offensive or polarizing content. The synthetic data generation process involved prompting LLMs to produce polarizing content for training purposes only. These synthetic samples are not representative of any real individuals or groups and are used solely for classifier training.

% TODO: Add acknowledgments for camera-ready
\section*{Acknowledgments}
This research was supported in part by Lilly Endowment, Inc., through its support for the Indiana University Pervasive Technology Institute.

\bibliography{references}

@inproceedings{mhalgi2024iucl,
    title = {{IUCL} at {PAN} 2024: Using Data Augmentation for Conspiracy Theory Detection},
    author = {Mhalgi, Shrirang and Pulipaka, Srikar Kashyap and K\"{u}bler, Sandra},
    booktitle = {Working Notes of CLEF 2024 -- Conference and Labs of the Evaluation Forum},
    year = {2024},
    address = {Grenoble, France}
}

@inproceedings{cegin2025llms,
    title = {{LLMs} vs Established Text Augmentation Techniques for Classification: When do the Benefits Outweigh the Costs?},
    author = {Cegin, Jan and Simko, Jakub and Brusilovsky, Peter},
    booktitle = {Proceedings of the 2025 Conference of the North American Chapter of the Association for Computational Linguistics (NAACL)},
    year = {2025},
    url = {https://aclanthology.org/2025.naacl-long.526/}
}

@article{yong2024lexcgen,
    title = {{LexC-Gen}: Generating Data for Extremely Low-Resource Languages with Large Language Models and Bilingual Lexicons},
    author = {Yong, Zheng Xin and Menghini, Cristina and Bach, Stephen H.},
    journal = {Computing Research Repository},
    volume = {arXiv:2402.14086},
    year = {2024},
    url = {https://arxiv.org/abs/2402.14086}
}

@inproceedings{sennrich2016improving,
    title = {Improving Neural Machine Translation Models with Monolingual Data},
    author = {Sennrich, Rico and Haddow, Barry and Birch, Alexandra},
    booktitle = {Proceedings of the 54th Annual Meeting of the Association for Computational Linguistics (Volume 1: Long Papers)},
    pages = {86--96},
    year = {2016},
    address = {Berlin, Germany},
    publisher = {Association for Computational Linguistics},
    doi = {10.18653/v1/P16-1009},
    url = {https://aclanthology.org/P16-1009/}
}

@inproceedings{edunov2018understanding,
    title = {Understanding Back-Translation at Scale},
    author = {Edunov, Sergey and Ott, Myle and Auli, Michael and Grangier, David},
    booktitle = {Proceedings of the 2018 Conference on Empirical Methods in Natural Language Processing},
    pages = {489--500},
    year = {2018},
    address = {Brussels, Belgium},
    publisher = {Association for Computational Linguistics},
    doi = {10.18653/v1/D18-1045},
    url = {https://aclanthology.org/D18-1045/}
}

@inproceedings{mostafazadeh-davani-etal-2021-improving,
    title = {Improving Counterfactual Generation for Fair Hate Speech Detection},
    author = {Mostafazadeh Davani, Aida and Omrani, Ali and Kennedy, Brendan and Atari, Mohammad and Ren, Xiang and Dehghani, Morteza},
    booktitle = {Proceedings of the 5th Workshop on Online Abuse and Harms (WOAH 2021)},
    pages = {92--101},
    year = {2021},
    address = {Online},
    publisher = {Association for Computational Linguistics},
    doi = {10.18653/v1/2021.woah-1.10},
    url = {https://aclanthology.org/2021.woah-1.10/}
}

@inproceedings{hu2021lora,
    title = {{LoRA}: Low-Rank Adaptation of Large Language Models},
    author = {Hu, Edward J. and Shen, Yelong and Wallis, Phillip and Allen-Zhu, Zeyuan and Li, Yuanzhi and Wang, Shean and Wang, Lu and Chen, Weizhu},
    booktitle = {Proceedings of the Tenth International Conference on Learning Representations (ICLR)},
    year = {2022},
    url = {https://arxiv.org/abs/2106.09685}
}

@article{lipton2014thresholding,
    title = {Thresholding Classifiers to Maximize {F1} Score},
    author = {Lipton, Zachary C. and Elkan, Charles and Narayanaswamy, Balakrishnan},
    journal = {Computing Research Repository},
    volume = {arXiv:1402.1892},
    year = {2014},
    url = {https://arxiv.org/abs/1402.1892}
}

@article{pillai2013threshold,
    title = {Threshold Optimisation for Multi-label Classifiers},
    author = {Pillai, Ignazio and Fumera, Giorgio and Roli, Fabio},
    journal = {Pattern Recognition},
    volume = {46},
    number = {7},
    pages = {2055--2065},
    year = {2013},
    publisher = {Elsevier},
    doi = {10.1016/j.patcog.2013.01.012}
}

@inproceedings{reimers2019sentence,
    title = {Sentence-{BERT}: Sentence Embeddings using Siamese {BERT}-Networks},
    author = {Reimers, Nils and Gurevych, Iryna},
    booktitle = {Proceedings of the 2019 Conference on Empirical Methods in Natural Language Processing and the 9th International Joint Conference on Natural Language Processing (EMNLP-IJCNLP)},
    pages = {3982--3992},
    year = {2019},
    address = {Hong Kong, China},
    publisher = {Association for Computational Linguistics},
    doi = {10.18653/v1/D19-1410},
    url = {https://aclanthology.org/D19-1410/}
}

@inproceedings{dettmers2023qlora,
    title = {{QLoRA}: Efficient Finetuning of Quantized Language Models},
    author = {Dettmers, Tim and Pagnoni, Artidoro and Holtzman, Ari and Zettlemoyer, Luke},
    booktitle = {Advances in Neural Information Processing Systems},
    volume = {36},
    year = {2023},
    url = {https://arxiv.org/abs/2305.14314}
}

@misc{gemma2024,
      title={Gemma: Open Models Based on Gemini Research and Technology}, 
      author={Gemma Team and Thomas Mesnard and Cassidy Hardin and Robert Dadashi and Surya Bhupatiraju and Shreya Pathak and Laurent Sifre and Morgane Rivière and Mihir Sanjay Kale and Juliette Love and Pouya Tafti and Léonard Hussenot and Pier Giuseppe Sessa and Aakanksha Chowdhery and Adam Roberts and Aditya Barua and Alex Botev and Alex Castro-Ros and Ambrose Slone and Amélie Héliou and Andrea Tacchetti and Anna Bulanova and Antonia Paterson and Beth Tsai and Bobak Shahriari and Charline Le Lan and Christopher A. Choquette-Choo and Clément Crepy and Daniel Cer and Daphne Ippolito and David Reid and Elena Buchatskaya and Eric Ni and Eric Noland and Geng Yan and George Tucker and George-Christian Muraru and Grigory Rozhdestvenskiy and Henryk Michalewski and Ian Tenney and Ivan Grishchenko and Jacob Austin and James Keeling and Jane Labanowski and Jean-Baptiste Lespiau and Jeff Stanway and Jenny Brennan and Jeremy Chen and Johan Ferret and Justin Chiu and Justin Mao-Jones and Katherine Lee and Kathy Yu and Katie Millican and Lars Lowe Sjoesund and Lisa Lee and Lucas Dixon and Machel Reid and Maciej Mikuła and Mateo Wirth and Michael Sharman and Nikolai Chinaev and Nithum Thain and Olivier Bachem and Oscar Chang and Oscar Wahltinez and Paige Bailey and Paul Michel and Petko Yotov and Rahma Chaabouni and Ramona Comanescu and Reena Jana and Rohan Anil and Ross McIlroy and Ruibo Liu and Ryan Mullins and Samuel L Smith and Sebastian Borgeaud and Sertan Girgin and Sholto Douglas and Shree Pandya and Siamak Shakeri and Soham De and Ted Klimenko and Tom Hennigan and Vlad Feinberg and Wojciech Stokowiec and Yu-hui Chen and Zafarali Ahmed and Zhitao Gong and Tris Warkentin and Ludovic Peran and Minh Giang and Clément Farabet and Oriol Vinyals and Jeff Dean and Koray Kavukcuoglu and Demis Hassabis and Zoubin Ghahramani and Douglas Eck and Joelle Barral and Fernando Pereira and Eli Collins and Armand Joulin and Noah Fiedel and Evan Senter and Alek Andreev and Kathleen Kenealy},
      year={2024},
      eprint={2403.08295},
      archivePrefix={arXiv},
      primaryClass={cs.CL},
      url={https://arxiv.org/abs/2403.08295}, 
}

@misc{naseem2026polarbenchmarkmultilingualmulticultural,

      title={POLAR: A Benchmark for Multilingual, Multicultural, and Multi-Event Online Polarization}, 

      author={Usman Naseem and Robert Geislinger and Juan Ren and Sarah Kohail and Rudy Garrido Veliz and P Sam Sahil and Yiran Zhang and Marco Antonio Stranisci and Idris Abdulmumin and Özge Alacam and Cengiz Acartürk and Aisha Jabr and Saba Anwar and Abinew Ali Ayele and Simona Frenda and Alessandra Teresa Cignarella and Elena Tutubalina and Oleg Rogov and Aung Kyaw Htet and Xintong Wang and Surendrabikram Thapa and Kritesh Rauniyar and Tanmoy Chakraborty and Arfeen Zeeshan and Dheeraj Kodati and Satya Keerthi and Sahar Moradizeyveh and Firoj Alam and Arid Hasan and Syed Ishtiaque Ahmed and Ye Kyaw Thu and Shantipriya Parida and Ihsan Ayyub Qazi and Lilian Wanzare and Nelson Odhiambo Onyango and Clemencia Siro and Jane Wanjiru Kimani and Ibrahim Said Ahmad and Adem Chanie Ali and Martin Semmann and Chris Biemann and Shamsuddeen Hassan Muhammad and Seid Muhie Yimam},

      year={2026},

      eprint={2505.20624},

      archivePrefix={arXiv},

      primaryClass={cs.CL},

      url={https://arxiv.org/abs/2505.20624}, 

}

@inproceedings{naseem-etal-2026-polar,

  title     = {{S}em{E}val-2026 Task 9: Detecting Multilingual, Multicultural and Multievent Online Polarization},

  author    = {Naseem, Usman and

               Geislinger, Robert and

               Ren, Juan and

               Kohail, Sarah and

               Garrido Veliz, Rudy and

               Sam Sahil, P and

               Zhang, Yiran and

               Stranisci, Marco Antonio and

               Abdulmumin, Idris and

               Alacam, {\"O}zge and

               Acart{\"u}rk, Cengiz and

               Jabr, Aisha and

               Anwar, Saba and

               Ayele, Abinew Ali and

               Tutubalina, Elena and

               Htet, Aung Kyaw and

               Wang, Xintong and

               Thapa, Surendrabikram and

               Chakraborty, Tanmoy and

               Kodati, Dheeraj and

               Moradizeyveh, Sahar and

               Alam, Firoj and

               Thu, Ye Kyaw and

               Parida, Shantipriya and

               Qazi, Ihsan Ayyub and

               Onyango, Nelson Odhiambo and

               Siro, Clemencia and

               Ahmad, Ibrahim Said and

               Wanzare, Lilian and

               Ali, Adem Chanie and

               Semmann, Martin and

               Biemann, Chris and

               Muhammad, Shamsuddeen Hassan and

               Yimam, Seid Muhie},

  booktitle = {Proceedings of the 20th International Workshop on Semantic Evaluation (SemEval-2026)},

  year      = {2026},

  publisher = {Association for Computational Linguistics},

}

\appendix

\section{Full Per-Language Results}
\label{sec:full_results}

Table~\ref{tab:full_results} shows complete test F1 scores for all 22 languages across our key submissions.

\begin{table*}[t]
\centering
\small
\begin{tabular}{lccccl}
\toprule
\textbf{Lang} & \textbf{Sub 1} & \textbf{Sub 4} & \textbf{$\Delta$} & \textbf{Strategy} & \textbf{Threshold} \\
\midrule
amh & 0.800 & 0.797 & $-$0.3\% & 27B tuned & 0.7 \\
arb & 0.848 & 0.848 & +0.0\% & 27B tuned & 0.4 \\
ben & 0.828 & 0.837 & +0.9\% & 27B tuned & 0.7 \\
deu & 0.721 & 0.728 & +0.7\% & Ens.\ weighted (0.3) & 0.45 \\
eng & 0.805 & 0.818 & +1.3\% & Ens.\ weighted (0.6) & 0.4 \\
fas & 0.804 & 0.828 & +2.4\% & Ens.\ weighted (0.6) & 0.6 \\
hau & 0.793 & 0.800 & +0.7\% & Ens.\ weighted (0.7) & 0.5 \\
hin & 0.800 & 0.824 & +2.4\% & Ens.\ average & 0.6 \\
ita & 0.543 & 0.563 & +2.0\% & Ens.\ weighted (0.6) & 0.45 \\
khm & 0.656 & 0.743 & +8.7\% & 12B tuned & 0.7 \\
mya & 0.874 & 0.877 & +0.3\% & 12B tuned & 0.35 \\
nep & 0.876 & 0.908 & +3.2\% & Ens.\ weighted (0.3) & 0.3 \\
ori & 0.793 & 0.811 & +1.8\% & Ens.\ weighted (0.7) & 0.4 \\
pan & 0.805 & 0.812 & +0.7\% & Ens.\ average & 0.45 \\
pol & 0.814 & 0.835 & +2.1\% & Ens.\ average & 0.3 \\
rus & 0.807 & 0.806 & $-$0.1\% & Ens.\ average & 0.55 \\
spa & 0.770 & 0.779 & +0.9\% & Ens.\ weighted (0.6) & 0.55 \\
swa & 0.774 & 0.811 & +3.7\% & Ens.\ average & 0.65 \\
tel & 0.893 & 0.882 & $-$1.1\% & 27B tuned & 0.5 \\
tur & 0.802 & 0.809 & +0.7\% & Ens.\ weighted (0.3) & 0.5 \\
urd & 0.803 & 0.803 & $-$0.0\% & 12B tuned & 0.35 \\
zho & 0.917 & 0.919 & +0.2\% & 12B tuned & 0.6 \\
\midrule
\textbf{Mean} & \textbf{0.797} & \textbf{0.811} & \textbf{+1.4\%} & & \\
\bottomrule
\end{tabular}
\caption{Complete per-language test F1 results with strategy and threshold used.}
\label{tab:full_results}
\end{table*}

\section{Synthetic Data Impact by Language}
\label{sec:synth_detail}

Table~\ref{tab:synth_per_lang} shows the optimal synthetic ratio for each language with the 12B model.

\begin{table}[t]
\centering
\small
\begin{tabular}{lccr}
\toprule
\textbf{Lang} & \textbf{Base} & \textbf{Best} & \textbf{Ratio} \\
\midrule
amh & 0.781 & 0.781 & 0\% \\
arb & 0.814 & 0.828 & 30\% \\
ben & 0.832 & 0.832 & 0\% \\
deu & -- & 0.792 & 30\% \\
eng & 0.808 & 0.811 & 50\% \\
fas & 0.847 & 0.849 & 30\% \\
hau & 0.766 & 0.804 & 20\% \\
hin & 0.848 & 0.865 & 30\% \\
ita & 0.686 & 0.701 & 50\% \\
khm & 0.628 & 0.692 & 30\% \\
mya & 0.853 & 0.887 & 20\% \\
nep & 0.870 & 0.890 & 10\% \\
ori & 0.891 & 0.891 & 0\% \\
pan & 0.859 & 0.859 & 0\% \\
pol & 0.792 & 0.834 & 30\% \\
rus & 0.781 & 0.827 & 30\% \\
spa & 0.757 & 0.757 & 0\% \\
swa & 0.819 & 0.833 & 10\% \\
tel & 0.907 & 0.907 & 0\% \\
tur & 0.809 & 0.817 & 10\% \\
urd & 0.785 & 0.799 & 10\% \\
zho & 0.921 & 0.930 & 30\% \\
\bottomrule
\end{tabular}
\caption{Optimal synthetic ratio per language for Gemma~3 12B. Base = 0\% synthetic. 6 out of 22 languages perform best without synthetic data.}
\label{tab:synth_per_lang}
\end{table}

\section{Prompt Strategies}
\label{sec:prompts}

We used GPT-4o-mini for all synthetic data generation, with different prompt templates per strategy. All prompts instruct the model to output only the generated text in the target language, without labels, explanations, or meta-commentary.

\paragraph{Direct Generation.} We provided the model with a definition of polarization and asked it to generate a social media post in the target language. For polarized samples, the prompt specifies that the post should target a specific group using vilifying, stereotyping, or dehumanizing language and show an us-vs-them mentality. For non-polarized samples, the prompt asks for a post that discusses the same types of topics in a neutral or balanced way, without vilifying anyone. Both prompts include a randomly sampled topic hint from five categories (political, ethnic/racial, religious, social class, international relations), each containing six specific topics. We used a temperature of 0.9 and a maximum of 250 tokens.

\paragraph{Paraphrasing.} We provided the model with a real training sample and asked it to rewrite it using different words and phrasing while preserving the exact same meaning, sentiment, and level of polarization. The prompt specifies that the output should be in natural target-language expressions and within 20\% of the original word count. We used a lower temperature of 0.7 to keep paraphrases close to the originals.

\paragraph{Contrastive Pairs.} We asked the model to generate two versions of a social media post about a given topic: one polarized (containing vilification, stereotyping, or an us-vs-them mentality) and one non-polarized (neutral, balanced viewpoints). Both versions are required to discuss the same topic, be similar in length (30 to 60 words), and feel authentic to the target language. The output uses a structured format (\texttt{POLARIZED:} and \texttt{NON\_POLARIZED:}) that we parse programmatically. We used a temperature of 0.8 and a maximum of 500 tokens.

\paragraph{Backtranslation and Cross-Lingual Transfer.} In exploratory experiments, we used Google Cloud Translate for backtranslation by translating source text to a pivot language (English by default, with multi-pivot using English, German, French, and Spanish) and then back to the source language. For cross-lingual transfer experiments, we translated training data from a related higher-resource language to the target language (e.g., Hindi to Nepali, Bengali to Odia).

\section{Leaderboard Rankings}
\label{sec:leaderboard}

Table~\ref{tab:leaderboard} shows our per-language rankings on the official Subtask~1 leaderboard.

\begin{table}[t]
\centering
\small
\begin{tabular}{lcccc}
\toprule
\textbf{Lang} & \textbf{Rank} & \textbf{Ours} & \textbf{Best} & \textbf{$\Delta$} \\
\midrule
\textbf{amh}\textsuperscript{\dag} & \textbf{1} & \textbf{0.800} & \textbf{0.800} & \textbf{0.000} \\
arb\textsuperscript{\dag} & 2 & 0.848 & 0.849 & $-$0.0004 \\
ben & 12 & 0.837 & 0.863 & $-$0.025 \\
deu\textsuperscript{\ddag} & 8 & 0.728 & 0.761 & $-$0.033 \\
eng\textsuperscript{\dag} & 3 & 0.818 & 0.825 & $-$0.008 \\
fas\textsuperscript{\ddag} & 4 & 0.828 & 0.835 & $-$0.007 \\
hau & 12 & 0.800 & 0.834 & $-$0.034 \\
\textbf{hin}\textsuperscript{\dag} & \textbf{1} & \textbf{0.824} & \textbf{0.824} & \textbf{0.000} \\
ita & 25 & 0.563 & 0.730 & $-$0.167 \\
khm\textsuperscript{\ddag} & 6 & 0.743 & 0.774 & $-$0.032 \\
mya\textsuperscript{\ddag} & 9 & 0.877 & 0.891 & $-$0.014 \\
nep & 12 & 0.908 & 0.924 & $-$0.016 \\
ori\textsuperscript{\ddag} & 5 & 0.811 & 0.826 & $-$0.015 \\
pan\textsuperscript{\dag} & 2 & 0.812 & 0.826 & $-$0.014 \\
pol\textsuperscript{\dag} & 3 & 0.835 & 0.843 & $-$0.008 \\
rus\textsuperscript{\ddag} & 6 & 0.807 & 0.830 & $-$0.024 \\
spa & 14 & 0.779 & 0.803 & $-$0.024 \\
\textbf{swa}\textsuperscript{\dag} & \textbf{1} & \textbf{0.811} & \textbf{0.811} & \textbf{0.000} \\
tel\textsuperscript{\ddag} & 4 & 0.893 & 0.905 & $-$0.012 \\
tur\textsuperscript{\dag} & 3 & 0.809 & 0.833 & $-$0.024 \\
urd\textsuperscript{\ddag} & 4 & 0.803 & 0.820 & $-$0.017 \\
zho\textsuperscript{\ddag} & 7 & 0.919 & 0.932 & $-$0.013 \\
\midrule
\textbf{Avg} & \textbf{2} & \textbf{0.811} & \textbf{0.818} & $\mathbf{-0.007}$ \\
\bottomrule
\end{tabular}
\caption{Per-language rankings on the official Subtask~1 leaderboard. $\Delta = \text{Ours} - \text{Best}$. \textsuperscript{\dag}Top 3, \textsuperscript{\ddag}Top 4--9. Bold = 1st place. Average Best score is computed over participants who participated in at least 11 of 22 languages.}
\label{tab:leaderboard}
\end{table}

\end{document}